\documentclass[11pt]{article}

\usepackage[final]{acl}
\usepackage{times}
\usepackage{latexsym}
\usepackage[T1]{fontenc}
\usepackage[utf8]{inputenc}
\usepackage{microtype}
\usepackage{graphicx}
\usepackage{booktabs}
\usepackage{multirow}
\usepackage{amsmath}
\usepackage{amssymb}
\usepackage{xcolor}
\usepackage{tikz}
\usepackage{url}
\usepackage{placeins}
\usetikzlibrary{arrows.meta,positioning,fit,backgrounds,calc,decorations.pathreplacing}

\newcommand{\axle}{\textsc{axle}}

\title{Activating the Annotator:\\
Operationalising Lived Experience in Data Annotation}

\author{Sheza Munir \\
  Department of Computer Science \\
  University of Toronto \\
  \texttt{sheza@cs.toronto.edu} \\\And
  Ramaravind Kommiya Mothilal \\
  Faculty of Information \\
  University of Toronto \\
  \texttt{ram.mothilal@mail.utoronto.ca} \\\AND
  Syed Ishtiaque Ahmed \\
  Department of Computer Science \\
  University of Toronto \\
  \texttt{ishtiaque@cs.toronto.edu}}

\begin{document}
\maketitle

\begin{abstract}
Perspectivist NLP has established that annotator positionality shapes labels
for subjective tasks, but the field still lacks methods for \emph{operationalising}
it.
Positionality is typically operationalised as a demographic form completed at
recruitment: a static attribute recorded once and treated as a covariate, on the
assumption that the knowledge it indexes is active when the annotator reads an
item. This paper treats that assumption as a design problem and asks whether lived
experience can instead be primed by the task itself. We introduce \textsc{axle}, a
pre-annotation stage in which a scenario vignette anchors a short, axis-guided
conversation that is synthesised into a first-person micro-narrative and returned to
the annotator before labelling begins. In a 2$\times$2 between-subjects
study (104 annotators, 30 items, 1{,}040 annotations) on ambiguous hate speech
spanning immigration, religion and gender, elicitation was associated with higher
identity resonance and higher sensitivity to harm. It also \emph{increased} label
entropy and \emph{lowered} inter-annotator agreement: the stage appears to surface
plurality rather than manufacture consensus, which distinguishes it from annotator
training, whose success criterion is convergence. 
A qualitative analysis of the rationales shows that elicitation shifts what annotators cite as grounds for their reading, from the literal text toward their own situated knowledge. Annotators not recruited for relevant lived experience invoke a lived-experience axis markedly more often under elicitation, and the axis they invoke is often not the one screened for. Positionality behaves like a state the interface either activates or leaves dormant.
\end{abstract}

\section{Introduction}

Subjective tasks such as hate speech and toxicity detection have no
annotator-independent gold standard. Disagreement in these tasks is patterned rather
than random, and tracks annotators' identities, values and social positions
\citep{waseem2016racist, sap2019risk, sap2022annotators, plank2022problem}. This
observation grounds the \emph{perspectivist} programme in NLP: the position that
divergent labels on such tasks are legitimate readings produced from different social
locations rather than errors to be corrected, and that the appropriate modelling
target is therefore the distribution of readings across a population rather than a
single consensus label \citep{basile2021toward, fleisig2024perspectivist, munir2026consensus}. Its
methods preserve that signal: release annotator-level labels
\citep{prabhakaran2021releasing2}, model raters individually
\citep{davani2022dealing, mokhberian2024capturing, orlikowski2025}, and treat label
distributions rather than majority votes as the training target
\citep{fornaciari2021beyond, uma2021learning}.

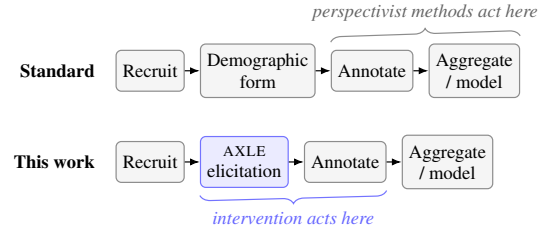
\begin{figure}[t]
\centering
\begin{tikzpicture}[
  font=\scriptsize,
  bx/.style={draw, rounded corners=2pt, minimum height=5.4mm, align=center,
             inner xsep=2.5pt, line width=.45pt, fill=black!4, draw=black!45},
  hl/.style={bx, fill=blue!8, draw=blue!55},
  a/.style={-{Latex[length=1.3mm]}, line width=.45pt},
  note/.style={font=\scriptsize\itshape, text=black!60}]
\node[bx] (r1) {Recruit};
\node[bx, right=2mm of r1] (d1) {Demographic\\form};
\node[bx, right=2mm of d1] (an1) {Annotate};
\node[bx, right=2mm of an1] (ag1) {Aggregate\\/ model};
\foreach \x/\y in {r1/d1, d1/an1, an1/ag1} \draw[a] (\x) -- (\y);
\node[left=1.4mm of r1, font=\scriptsize\bfseries, align=right] {Standard};
\draw[decorate, decoration={brace, amplitude=1.4mm}, draw=black!45, line width=.45pt]
  ([yshift=1mm]an1.north west) -- ([yshift=1mm]ag1.north east)
  node[midway, above=1mm, note] {perspectivist methods act here};
\node[bx, below=6.5mm of r1] (r2) {Recruit};
\node[hl, right=2mm of r2] (e2) {\axle{}\\elicitation};
\node[bx, right=2mm of e2] (an2) {Annotate};
\node[bx, right=2mm of an2] (ag2) {Aggregate\\/ model};
\foreach \x/\y in {r2/e2, e2/an2, an2/ag2} \draw[a] (\x) -- (\y);
\node[left=1.4mm of r2, font=\scriptsize\bfseries, align=right] {This work};
\draw[decorate, decoration={brace, mirror, amplitude=1.4mm}, draw=blue!55, line width=.45pt]
  ([yshift=-1mm]e2.south west) -- ([yshift=-1mm]an2.south east)
  node[midway, below=1mm, note, text=blue!60] {intervention acts here};
\end{tikzpicture}
\caption{Perspectivist methods operate
\emph{downstream}, preserving variation the annotation step already produced.
\axle{} operates \emph{upstream}, on how much variation that step produces at all.}
\label{fig:pipeline}
\end{figure}

This infrastructure is built almost entirely downstream (Figure~\ref{fig:pipeline}).
It assumes that whatever positional signal exists has already been inscribed by the
time labels are aggregated, and asks only how to preserve, model or exploit it. The
upstream question i.e., what determines how much positional variation the annotation
process generates in the first place, is rarely posed. In practice, positionality is
operationalised as a recruitment-time demographic form: checkboxes recorded once,
frozen, used as a covariate. \citet{fleisig2024perspectivist} note that demographic
attributes predict individual annotation behaviour poorly, and
\citet{orlikowski2025} show that conditioning on an annotator's own prior judgements
outperforms conditioning on their demographics by a wide margin; persona prompting
recovers a similarly small fraction of real annotator variance
\citep{hu2024quantifying, beck2024sensitivity}. A demographic category names a
position without making it operative.

Meanwhile the interface works against positional reasoning. Guidelines instruct
annotators to apply a definition consistently (the \emph{prescriptive} paradigm of
\citet{rottger2022contrasting}) and even nominally descriptive tasks are framed as
classification: read the item, pick a category, move on. That framing recruits a fast
rule-application posture in which an annotator's own history registers as bias to be
suppressed rather than as the evidence the task requires, and crowdwork economics
sharpen the incentive \citep{miceli2020, sambasivan2021}. Where lived experience is
what separates one reading of an ambiguous statement from another, this posture
favours the reading available to a person with no stake in the content.

The design question is therefore whether relevant experience can be brought into
working memory \emph{before} annotation begins. Interpretive phenomenology holds that
lived experience becomes available through narrative and interpretation, and that a
reflective register must be established before evaluative work
\citep{frechette2020capturing}; micro-narrative methods scale that elicitation to
large samples \citep{skeggs2025}. We adapt both into a pre-annotation stage,
\axle{} (Axis-guided eXperience ELicitation). Annotators read a scenario vignette
\citep{skilling2020vignettes, barter1999vignettes} thematically matched to the items
they will later label; a conversational elicitor probes their relationship to those
themes, structured invisibly around seven dimensions of lived experience
\citep{gautam2025, jacobson2019}; the conversation is synthesised into a first-person
micro-narrative that is returned to the annotator and stays accessible while they
work. Elicitation first establishes a narrative
register; annotation first would consolidate classification posture before any
positional material could surface.

We evaluate this in a preliminary 2$\times$2 between-subjects study on Prolific
($N{=}104$; 1{,}040 annotations) crossing elicitation against a matched control, and
lived-experience (LE) against non-lived-experience (non-LE) screeners, over 30
ambiguous hate speech items in three scenarios.

\subsection{Contributions}

\begin{itemize}\itemsep1pt
\item \textbf{\axle{}}, a pre-annotation lived-experience elicitation procedure, comprising a vignette, an axis-guided conversation and fidelity-constrained micro-narrative synthesis, released as runnable code. Against a matched no-elicitation arm it raised
identity resonance and harm sensitivity at both item and participant level.

\item \textbf{Evidence that \axle{} increases rather than collapses label
plurality}: per-item entropy rose and inter-annotator agreement fell---the opposite of what annotator training produces, and precisely what a perspectivist objective requires.

\item \textbf{A mechanism and a boundary condition.} Analysis of 240
rationales shows elicitation shifting what annotators cite as grounds, from the
literal text toward their own situated knowledge. Among annotators not screened for relevant lived experience, rationales invoke a lived-experience axis substantially more often under elicitation (a difference of roughly 38 percentage points in a subsample of 24 participants), frequently on an axis the screener never tested.
Positionality behaves like an activatable state rather than a fixed attribute.
\end{itemize}

Study materials, data and analysis code are available at
\url{https://github.com/ShezaMunir/vdda}.

\section{Related Work}

\paragraph{Preserving label variation.}
\citet{pavlick2019inherent} showed that disagreement in natural language inference is
reproducible rather than noisy, and \citet{plank2022problem} consolidated such
findings under \emph{human label variation}. Methods now range from soft-label
multi-task learning \citep{fornaciari2021beyond} and multi-annotator architectures
\citep{davani2022dealing} to annotator embeddings \citep{mokhberian2024capturing} and
jury-based adjudication \citep{Gordon2022}; \citet{Gordon2021} quantify how much
standard metrics overstate performance when variation is ignored. This work is
uniformly \emph{post hoc}: it asks how to preserve variation already produced, not
what determines how much of it the annotation process generates.

\paragraph{The proxy problem.}
Identity effects on labels are well documented \citep{waseem2016racist, sap2019risk,
Goyal2022}, and \citet{Sachdeva2022} formalise ``identity sensitivity'' using item
response theory. The recurrent operationalisation is demographic, and fails twice
over: categories are coarse relative to the constructs they proxy, and recording a
category does not ensure the knowledge it indexes is \emph{engaged} while the
annotator reads. That is the gap addressed here.

\paragraph{Eliciting lived experience.}
Methods for surfacing situated knowledge are well developed outside NLP. Vignettes
elicit value-laden constructs without asking participants to theorise about
themselves \citep{barter1999vignettes, skilling2020vignettes};
\citet{willig2013introducing} specifies an interview progression from initial
reaction through a stakes cue to deepening on one thread; \citet{skeggs2025} show
that fidelity-constrained micro-narratives scale. \citet{gautam2025} supply the axis
framework used here, and \citet{haraway1988situated} and \citet{bardzell2010feminist}
the epistemology. These methods are entering annotation research
\citep{tang2026disability, Lameiro2025, Phutane2025}, but as recruitment or co-design
practices, not as annotation-time procedures.

\paragraph{Reflexivity as an intervention.}
\citet{arzberger2026label} introduce \emph{reflexive annotating}, a probe attaching
identity facets and free-form rationales alongside each label, and report from a
qualitative study ($N{=}30$) that it elicits intersectional reasoning and positional
humility. Their probe adapts the Social Identity Map \citep{jacobson2019}: a static,
three-tier form administered concurrently with labelling. The present design differs
in position, instrument and evidence. Elicitation \emph{precedes} annotation, on the
premise that reflective register must be established before evaluation; it uses an
adaptive conversation producing an artefact returned to the annotator rather than a
fixed form; and it is evaluated against a matched control arm. Adaptivity also
addresses scale: a static map must be interpreted by each worker unaided and its yield
assessed through researcher-led interviews, whereas a conversational elicitor adapts
probes to what each participant has said and runs without researcher presence, which
is what makes per-annotator elicitation feasible at crowdsourcing volumes.

\section{Study Design}
\label{sec:method}

\begin{figure*}[t]
\centering
\begin{tikzpicture}[
  font=\small,
  stage/.style={draw, rounded corners=2pt, minimum height=8.5mm, minimum width=22mm,
                align=center, inner sep=3pt, line width=.5pt},
  elic/.style={stage, fill=blue!7, draw=blue!55},
  shared/.style={stage, fill=black!5, draw=black!45},
  arr/.style={-{Latex[length=1.6mm]}, line width=.5pt},
  lbl/.style={font=\scriptsize\itshape, text=black!65}]
\node[shared] (d1) {Disclosure};
\node[elic, right=4mm of d1] (v1) {Vignette};
\node[elic, right=4mm of v1] (c1) {\axle{} interview\\\scriptsize(max 5 turns)};
\node[elic, right=4mm of c1] (m1) {Micro-narrative\\\scriptsize(synthesis)};
\node[shared, right=4mm of m1] (a1) {Annotation\\\scriptsize(10 items)};
\node[shared, right=4mm of a1] (r1) {Reflexivity};
\foreach \a/\b in {d1/v1, v1/c1, c1/m1, m1/a1, a1/r1} \draw[arr] (\a) -- (\b);
\node[shared, below=13mm of d1] (a2) {Annotation\\\scriptsize(10 items)};
\node[shared, right=4mm of a2] (d2) {Disclosure};
\node[shared, right=4mm of d2] (r2) {Reflexivity};
\foreach \a/\b in {a2/d2, d2/r2} \draw[arr] (\a) -- (\b);
\node[left=3mm of d1, align=right, font=\small\bfseries] {Elicitation\\arm};
\node[left=3mm of a2, align=right, font=\small\bfseries] {Control\\arm};
\draw[arr, dashed, draw=blue!55] (m1.south) .. controls +(0,-6mm) and +(0,-6mm) .. (a1.south)
  node[midway, below, lbl, text=blue!55] {narrative available during labelling};
\end{tikzpicture}
\caption{Stage order in the two arms. Both share the disclosure, annotation and
reflexivity stages and use an identical label taxonomy, rationale prompt and
resonance item. In the control arm the three shaded stages are removed and disclosure
runs \emph{after} annotation, so that no identity-related framing precedes labelling.}
\label{fig:design}
\end{figure*}
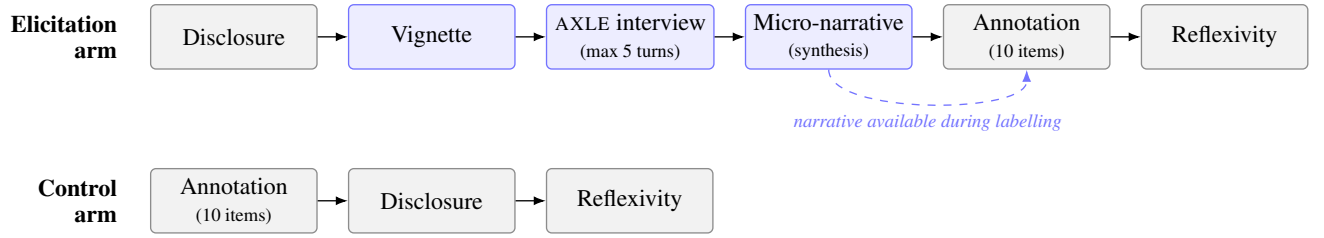

\subsection{Task and Items}

Annotators label short social-media-style posts on a five-point ordinal harm scale
(\emph{not harmful}, \emph{likely not harmful}, \emph{ambiguous}, \emph{likely
harmful}, \emph{clearly harmful}), select a target category, write a rationale
(minimum 50 words), and rate how relevant their identity or personal experience felt
to the item on a 1--5 scale, termed \emph{identity resonance}.

Items come from the ambiguous hate speech corpus we constructed under the codebook
developed for this study, requiring all four of the following. \textbf{C1 (surface deniability):} the statement reads as
policy concern, cultural commentary or empirical claim, and a reader unfamiliar with
the relevant discourse would not flag it on first pass. \textbf{C2 (experiential
divergence):} a reader with direct lived experience of the targeted group would
plausibly read it differently in propositional content or implication, not merely
more sensitively. \textbf{C3 (discourse embeddedness):} it borrows framing or
rhetorical structure circulating in exclusionary discourse communities, legible as
such to those inside it. \textbf{C4 (no explicit derogation):} no slurs, calls for
harm or direct dehumanisation.

C2 is the operative constraint: items where disagreement would follow only from
lexical vagueness, or where toxicity is near-universal across backgrounds, are
excluded. The corpus is therefore not a random sample of hate speech but a
concentration of the hardest cases (those where an annotator's background is most
likely to change the reading), so effect sizes here should be read as an upper bound
on what elicitation would achieve on a general corpus.

Items are organised into three thematic scenarios of ten items each: A
(immigration/nativism), B (religion), C (gender/sexuality), and tagged with a
primary domain: \textsc{imm}, \textsc{rel}, \textsc{gen}, or \textsc{int}
(intersectional). Scenarios B and C each contain four intersectional items. Items are
synthetic, generated against the codebook and screened by the authors reading each
candidate against C1--C4, and the criteria
were validated in a pilot round reported in the code repository. The full corpus is
in Appendix~\ref{app:items}.

\subsection{Conditions}

\paragraph{Elicitation.} The elicitation arm completes the full instrument in
Figure~\ref{fig:design}. The control arm is the same application with the vignette,
interview and synthesis stages removed, disclosure relocated after annotation, no
mention of positionality or lived experience in the study introduction, and no
narrative panel in the annotation interface. Label taxonomy, target options, rationale
prompt, word minimum, resonance item and reflexivity stage are identical across arms.
Both arms are descriptive-paradigm annotation in the sense of
\citet{rottger2022contrasting}; the contrast is latent versus activated
positionality, not descriptive versus prescriptive framing. 

\paragraph{Recruitment.} Within each scenario, two strata are recruited using Prolific
pre-screeners selecting for and against direct experience of the relevant axis: for
scenario A, migration status; for B, minority religious affiliation; for C, gender identity and sexual orientation. Full screener
specifications are in Appendix~\ref{app:screeners}. All participants are US or
Canadian residents, fluent in English, with $\geq$95\% approval, and are excluded
from concurrent studies in the project.

\subsection{The Elicitation Instrument}

\paragraph{Disclosure.} Participants report their connection to the topic (direct
personal experience / close community member / professional or academic / interested
observer), its duration, and a short description---an independent check on the
screeners (\S\ref{sec:participants}).

\paragraph{Vignette-anchored interview.} Participants read an 80--150 word vignette
depicting a contested public situation matched to their scenario---a town hall over
migrant housing (A), a boycott of a bakery celebrating a minority religious holiday
(B), a campus dispute over an op-ed on identity politics (C)---constructed following
\citet{barter1999vignettes} and \citet{skilling2020vignettes}. Vignettes describe a
situation without taking a position, so that any stance the participant adopts
originates with them.

\axle{} then conducts a conversation seeded with a fixed opening question. Its system
prompt encodes seven lived-experience axes \citep{gautam2025}: sociocultural and
geographic context; linguistic background and dialect; socioeconomic status and
labour; race and ethnicity; gender identity and sexuality; disability and
neurodivergence; and epistemic proximity to those most affected. The axes function as
invisible lenses; the model is instructed never to name or enumerate them. Coverage
is tracked with a keyword heuristic over participant turns, and each subsequent
prompt is constructed to bridge from the participant's last utterance to the nearest
uncovered axis, implementing the hermeneutic principle that later prompts should be
grounded in earlier answers \citep{frechette2020capturing}.

Turn policy follows \citet{willig2013introducing}: turn 1 anchors on the most specific
or emotionally loaded element of the first response, turns 2--3 broaden to uncovered
axes, turn 4 integrates, and turn 5 is a hard cap with a graceful wind-down. A
\emph{repair} rule overrides broadening. If the last response is under 12 words,
\axle{} stays on the thread and asks for a concrete memory. Every turn is constrained
to one question, one to two sentences, no affirmations, no paraphrase, and no generic
questions. The flow was fixed after pilot rounds; earlier prompt versions are in the
code repository and the final prompts in Appendix~\ref{app:prompts}.

\paragraph{Micro-narrative synthesis.} The transcript is synthesised into a
first-person narrative of at least 120 words under fidelity-first constraints adapted
from \citet{skeggs2025}: preserve the participant's own words, add nothing they did
not express, cover at least three distinct dimensions of lived experience. The
narrative is shown to the participant for review and remains accessible in a
collapsible panel throughout annotation.

\paragraph{Annotation.} Ten items are presented one at a time in fixed order, with
per-item timing recorded from first render to submission; all four responses are
required. The rationale prompt invites participants to report whether any aspect of
their identity, beliefs, values or lived experience affected their reading. The
resonance item is framed as identity resonance rather than bias, avoiding language
that frames disclosure as an admission of error.

\paragraph{Disagreement-anchored reflexivity.}
\label{sec:reflex}
After annotation, the system retrieves completed sessions from other participants in
the same scenario and builds up to three \emph{disagreement cards}: the post text, up
to three anonymised other participants' labels and rationales, and the participant's
own label and rationale, shown last. Participants then write a minimum-80-word
reflection on what in their background shaped a reading that differed from others'.
This stage runs in both arms.

\paragraph{Models.} \axle{} and the synthesiser use \textsc{Qwen2.5-7B-Instruct} via
the HuggingFace Inference API and \textsc{Qwen3-32B} via the Groq API
\citep{qwen2024qwen25, qwen2025qwen3}, temperature 0.3, with reasoning traces
stripped. Conversations reached a mean of 4.88 participant turns ($SD{=}0.32$),
nearly all running to the five-turn cap rather than triggering the turn-4 stopping
condition, and micro-narratives averaged 174 words ($SD{=}54.7$).

\subsection{Participants}
\label{sec:participants}

104 annotators were recruited via Prolific \citep{palan2018prolific}, paid
CAD~15--20/hour prorated by condition. Table~\ref{tab:cells} gives the design cells. Scenario A was fielded first and carries
a larger sample because it doubled as the instrument pilot: the turn policy, repair
rule and synthesis constraints were finalised on scenario A data before B and C were
launched. Participants averaged 36.0
years ($SD{=}10.9$, range 18--65), 56 male and 48 female, resident in Canada (59) or
the US (45), born in 23 countries, with 38 fluent in more than one language; 49 had
migrated to their country of residence. Full demographics are in
Appendix~\ref{app:demographics}.

The disclosure stage checks the screeners: stratum and self-reported connection agree
for 97 of 104 participants (93.3\%). All 52 LE participants report direct personal
experience; the seven mismatches are non-LE participants who nonetheless report it,
concentrated in scenarios B (5 of 12) and C (2 of 12), where the screeners test a
narrower axis than the disclosure question.

Three measures discouraged model-generated
rationales: a disclaimer asking for the participant's own words, disabled pasting in
free-text fields, and Prolific's submission-speed screening.

\begin{table}[t]
\centering\small
\begin{tabular}{lcccc}
\toprule
& \multicolumn{3}{c}{Scenario} & \\
\cmidrule(lr){2-4}
Condition & A & B & C & Total \\
\midrule
LE--elicit        & 18 & 6 & 6 & 30 \\
LE--control       & 10 & 6 & 6 & 22 \\
non-LE--elicit    & 18 & 6 & 6 & 30 \\
non-LE--control   & 10 & 6 & 6 & 22 \\
\midrule
Total             & 56 & 24 & 24 & 104 \\
\bottomrule
\end{tabular}
\caption{Design cells. A = immigration, B = religion, C = gender/sexuality. Each
participant annotates all ten items in their scenario (1{,}040 annotations).}
\label{tab:cells}
\end{table}

\subsection{Measures and Analysis}

Primary measures are identity resonance and harm rating (the five-point label mapped
to an ordinal 1--5 scale); secondary measures are per-item time, label entropy and
target attribution. Both scales are ordinal and non-normal, so two-sided
Mann--Whitney $U$ tests are used with the rank-biserial correlation $r$ as effect
size. Annotations are nested within participants, so item-level tests violate
independence; \textbf{participant-level tests are treated as primary} ($n{=}60$
vs.\ $44$, one mean per participant), with item-level tests ($n{=}600$ vs.\ $440$)
secondary. Per-item comparisons use paired Wilcoxon signed-rank tests over the 30
items, and agreement is reported as ordinal Krippendorff's $\alpha$
\citep{krippendorff2004content} computed within condition.

\section{Results}
\label{sec:results}

\subsection{Resonance and Harm Sensitivity}

\begin{figure*}[t]
\centering
\includegraphics[width=.37\textwidth]{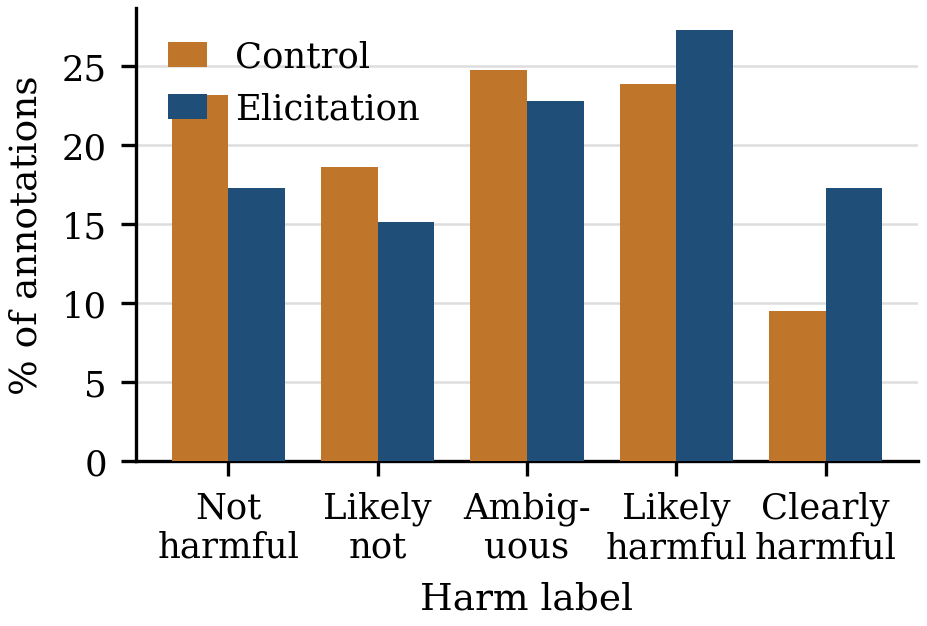}\hfill
\includegraphics[width=.37\textwidth]{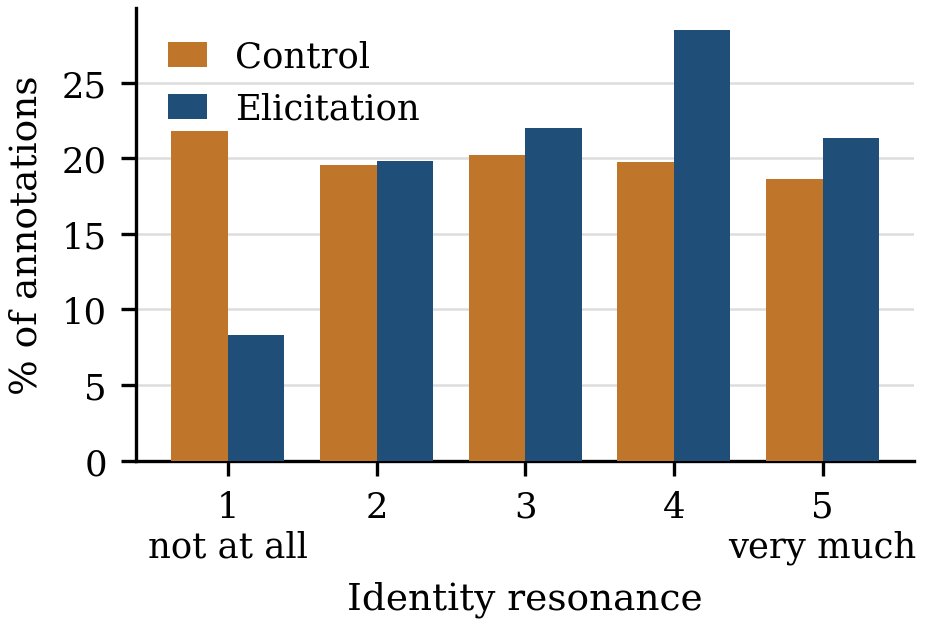}
\caption{Response distributions by arm. Left: harm labels; elicitation moves mass
toward both ends, with the largest single change at \emph{clearly harmful}. Right:
identity resonance; the proportion of items rated ``not at all relevant to my
identity or experience'' falls by more than half.}
\label{fig:dists}
\end{figure*}

Elicitation raised identity resonance ($M{=}3.35$ vs.\ $2.94$; participant level
$p{=}.041$, $r{=}.24$; item level $p{=}3.6\times10^{-6}$) and harm ratings
($M{=}3.12$ vs.\ $2.78$; participant level $p{=}.029$, $r{=}.25$; item level
$p{=}3.9\times10^{-5}$). Appendix~\ref{app:stats} (Table~\ref{tab:main}) summarises. Both shifts concentrate at the poles: annotations rated 1 on
resonance fall from 21.8\% to 8.3\%, and \emph{clearly harmful} judgements rise from
9.5\% to 17.3\% while \emph{not harmful} falls from 23.2\% to 17.3\%
(Figure~\ref{fig:dists}). Twenty-five of 30 items are rated more resonant and 25 of 30 more harmful under elicitation, with 20 items higher on both.

The upward harm shift admits two readings. Because every item satisfies C1--C4, it is
consistent with elicitation surfacing previously unavailable readings; it is equally
consistent with a demand characteristic. The entropy analysis below is the primary
evidence against the second.

\subsection{Elicitation Increases Plurality}
\label{sec:entropy}

\begin{figure}[t]
\centering
\includegraphics[width=.78\columnwidth]{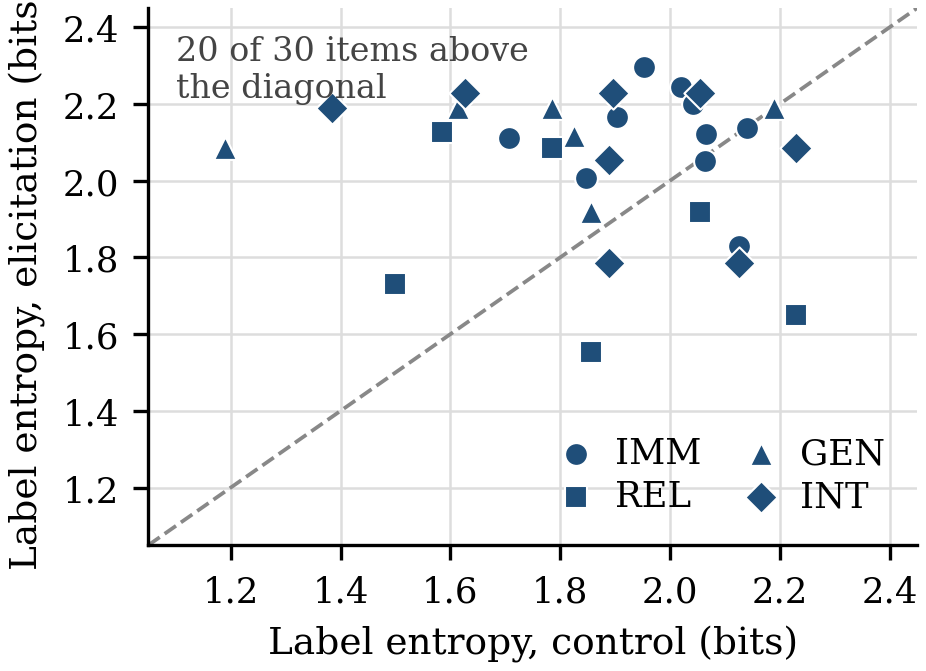}
\caption{Per-item Shannon entropy of the harm label distribution. Points above the
diagonal are items on which elicitation produced more dispersed labels.}
\label{fig:entropy}
\end{figure}

If elicitation simply trained annotators toward a shared reading, agreement would
rise. It falls. Mean per-item Shannon entropy of the five-point harm distribution was
$2.05$ bits under elicitation versus $1.88$ under control (Wilcoxon $W{=}100$,
$p{=}.011$; 20 of 30 items higher; Figure~\ref{fig:entropy}). Pooled ordinal
Krippendorff's $\alpha$ was $.066$ for elicitation and $.111$ for control, with the
same direction in scenarios B and C (Appendix~\ref{app:stats}).

These $\alpha$ values are very low and should not be read as reliability estimates: a
corpus built to maximise experiential divergence is one on which high agreement is
definitionally unattainable, so near-chance $\alpha$ is a property of the sampling
frame rather than a defect of the annotators. Only the between-arm comparison at a
fixed item set is interpretable, and it agrees with the entropy result.

Interventions that raise annotator engagement usually raise agreement, because they
converge annotators on the guideline's intended reading. Elicitation raises
engagement while lowering agreement: it recruits different readings for different
people rather than one better-trained reading. Under a prescriptive objective this is
a failure mode; under a perspectivist objective, where the target is the distribution
of valid readings \citep{basile2021toward, plank2022problem}, it is the intended
behaviour, and it is what separates elicitation from annotator training. Inattention would also raise dispersion, but would shorten rationales and reduce time on item; neither occurs (Appendix~\ref{app:stats}).

\subsection{Effects Are Larger for Non-LE Annotators}
\label{sec:strata}

Recruitment stratum alone had a limited effect: LE annotators reported higher
resonance than non-LE annotators ($3.38$ vs.\ $2.97$, $p{=}.026$) but did not differ
in harm ratings ($p{=}.61$). Screening for lived experience changed how relevant
annotators felt the content was without changing what they labelled.

Elicitation effects appeared in both strata but were larger among non-LE annotators.
At item level, resonance rose $0.51$ points for non-LE ($3.19$ vs.\ $2.68$,
$p{=}1.1\times10^{-5}$, $r{=}.22$) against $0.31$ for LE ($p{=}.023$, $r{=}.11$);
harm rose $0.40$ for non-LE ($p{=}.0008$) against $0.28$ for LE ($p{=}.022$).
Elicited non-LE annotators ($3.19$) reported resonance level on a level with unelicited LE
annotators ($3.20$). Section~\ref{sec:qual} identifies what they were drawing on.

\subsection{Domain Effects}

Elicitation raised identity resonance in every domain, but the effect on harm ratings
took two different forms (Appendix~\ref{app:domain}). For \textsc{rel} and \textsc{int}
items it was a uniform shift: mean harm rose ($2.53\to3.33$ and $2.91\to3.27$) while
label dispersion stayed roughly constant, so annotators moved together toward a harsher
reading. For \textsc{gen} items the mean did not move at all ($2.94\to2.88$), yet
resonance rose sharply ($2.54\to3.28$) and entropy rose more than in any other domain
($1.74\to2.11$). Annotators engaged with these items more, and moved in opposing
directions: the movement cancels in the mean and appears in the variance. Domain
cells are small ($n{=}6$--$10$ items), so we report this as descriptive.

\subsection{Cost}
\label{sec:cost}

What elicitation adds is a fixed cost: median session time was 51.4~min against
43.8~min ($p{=}.010$), and median non-annotation overhead 18.6~min against 7.9~min.
Because the marginal per-item cost is flat or slightly negative, that overhead
amortises---extrapolating from medians, projected cumulative cost crosses at roughly
48 items (Appendix~\ref{app:domain}). The projection is illustrative rather than
measured, resting on a per-item advantage that is not significant and ignoring
fatigue.

\section{Qualitative Analysis}
\label{sec:qual}

The quantitative results establish that elicitation changes labels and raises
dispersion, but not what annotators bring to bear when it does. A reflexive thematic
analysis \citep{braun2022thematic} of the free-text corpus addresses that.

\subsection{Sample and Protocol}

The analysis covers 276 units from a stratified subsample of 24 participants (six per
design cell, two per scenario) comprising 240 rationales, 12 elicitation transcripts
and 24 reflexivity responses. Two coders coded the same units independently, then compared codes; disagreements were
resolved by discussion and definitions revised. Statistics use the reconciled codebook
(Appendix~\ref{app:codebook}). Agreement on the construct carrying the central claim
below (whether a rationale invokes situated grounding) was near-perfect
($\kappa{=}.84$); per-construct values are in Appendix~\ref{app:kappa} (Figure~\ref{fig:kappa}).

\begin{figure*}[t]
\centering
\includegraphics[width=.5\textwidth]{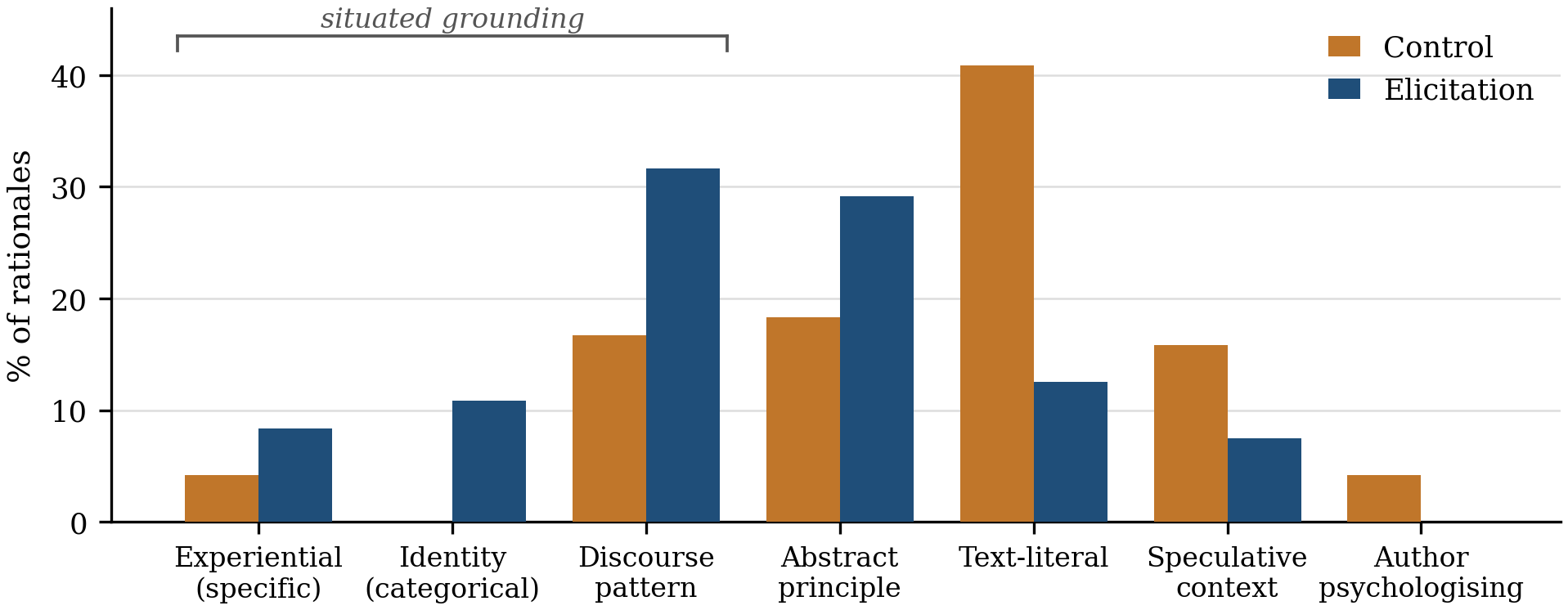}\hspace{2cm}
\includegraphics[width=.22\textwidth]{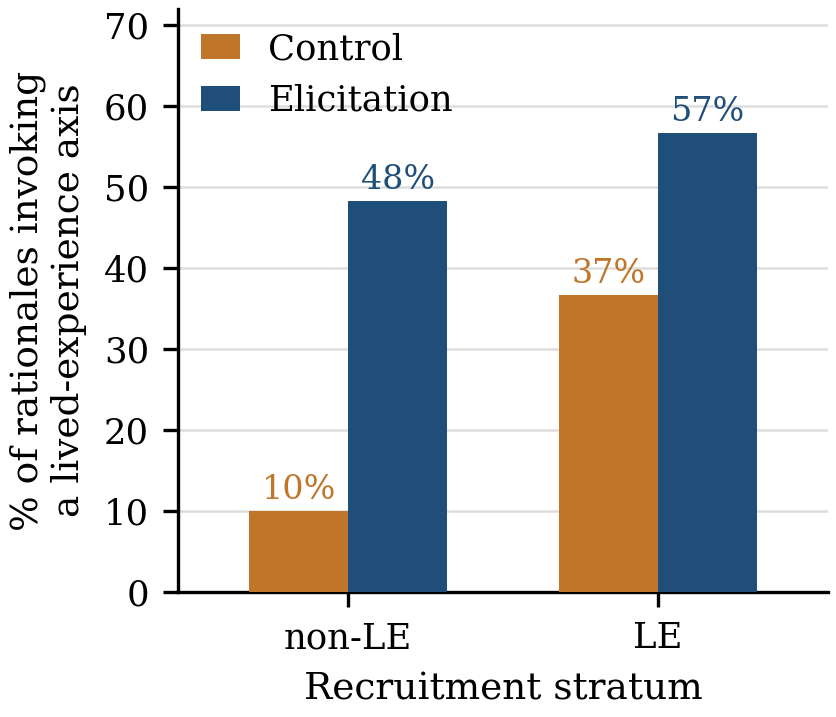}
\caption{Left: what the rationale treats as grounds for its reading, by arm
($n{=}240$); elicitation moves mass out of literal reading and into situated
grounding. Right: proportion of rationales invoking any of the seven lived-experience
axes, by design cell; the elicitation gain is larger in the non-LE stratum.}
\label{fig:qual}
\end{figure*}
\subsection{Elicitation Shifts the Grounds of Judgement}

Every rationale states, explicitly or implicitly, what the annotator treats as
evidence for their reading. We coded that: is the judgement grounded in a personal
episode, a claim of group membership, a recognised discourse pattern, an abstract
principle, the literal wording, guessed-at context, or the author's presumed state of
mind? The first three are grouped as \emph{situated grounding}, since each asserts
that the annotator's own position bears on the reading.

Situated grounding accounts for 50.8\% of elicitation-arm rationales against 20.8\%
under control (Fisher $p{=}1.9\times10^{-6}$; participant-level $p{=}.041$), while
appeals to the literal wording alone fall from 40.8\% to 12.5\%
(Figure~\ref{fig:qual}, left); claims of group membership are absent from the control
arm entirely.

The stance annotators take, by contrast, barely moves. We coded whether each rationale
commits to a single reading, names two or more readings attributable to different
readers, or withholds judgement for want of context. Commitment to a single reading
accounts for 64.2\% of control and 70.8\% of elicitation rationales, and acknowledged
multiplicity is identical at 16.7\%. Elicitation does not make annotators more
tentative; it makes them differently grounded while remaining equally committed, which
is the combination that produces the entropy result. Annotators are not converging on
uncertainty but diverging on confident readings reached from different starting points.

The arm-level figure averages over a bimodal distribution: grounding shifts on items
where the micro-narrative surfaces in the rationale and not on items where it does
not, a pattern that holds within participants rather than between them
(Appendix~\ref{app:kappa}).

\subsection{What Non-LE Annotators Draw On}

Section~\ref{sec:strata} left open why elicitation effects were larger among
annotators not screened for relevant lived experience. Coding which of the seven axes
each rationale invokes answers it (Figure~\ref{fig:qual}, right). Among non-LE
annotators, axis invocation rises from 10.0\% under control to 48.3\% under
elicitation, against 36.7\% to 56.7\% among LE annotators (participant-level
$p{=}.027$). The gain is carried largely by the sociocultural and geographic
axis, which rises from 4.2\% to 25.8\% of all rationales.

The screeners test one axis per scenario while the instrument probes seven, and the
axis invoked is frequently not the screened one: non-LE participants in scenario C
ground readings in class and regional upbringing rather than gender. Elicitation
appears to locate whichever axis is live for a given annotator, which a single-axis
screener cannot do.

This does not make elicitation uniformly harm-amplifying. Several annotators ground a
\emph{lower} harm rating in their own position, offering the \emph{absence} of
familiarity with a discourse as the reason a benign reading is correct. In the qualitative analysis, lived
experience raises the variance of readings, not their mean severity: the qualitative
counterpart of the entropy finding.

\section{Discussion}

\paragraph{Positionality as a state.}
The dominant operationalisation in NLP is attributive: an annotator \emph{has}
demographic properties, recorded once, treated as stable. These results fit a
state-based account better. Screening for lived experience raised resonance but not
harm ratings; activating it raised both, and the largest gains in axis invocation
appeared in the stratum not screened for it---anomalous under an attributive account,
expected under a state account in which every annotator occupies a position on all
seven axes and what varies is whether any is engaged. That grounding shifts primarily where the narrative is live at the point of judgement
is what a state should do and what an attribute cannot. It would also help explain why annotator-history conditioning beats
demographic conditioning \citep{orlikowski2025} and why persona prompts recover so
little annotator variance \citep{hu2024quantifying}: both model the attribute while
the interface governs the state.

\paragraph{Plurality as the evaluation criterion.}
Annotation interventions are conventionally judged by whether they raise agreement,
and this one fails that test. If the target is the distribution rather than the mode
\citep{plank2022problem, basile2021toward}, the test is wrong: what matters is whether
an intervention reshapes the label distribution to better reflect the population of
readers.

\paragraph{Implications for crowdsourced annotation.}
Elicitation is a fixed per-worker cost with no per-item cost, which places it
structurally where qualification tests already sit on major crowdsourcing
platforms: expensive per worker, negligible per label once batch sizes are realistic.
A platform could run it once per worker per domain and reuse the narrative across
subsequent tasks, amortising the overhead across hundreds of items rather than the ten
measured here. What that reuse requires is coverage: whether a worker completed an
elicitation stage is the wrong quality signal, and whether its output is live when
they judge a given item is the right one. Interfaces that keep the narrative visible,
surface it against thematically matched items, or prompt against it where it has not
been invoked all target that directly, as does monitoring per-worker repetition rate,
which is cheap to compute and separates engagement from its appearance. Two further
consequences follow. Elicitation artefacts are annotator-level, text-valued metadata
and a natural conditioning signal for annotator-aware models
\citep{davani2022dealing}; unlike demographic vectors they are self-authored and not
confined to the categories a screener happened to test. Agreement-based quality assurance penalises the very signal a perspectivist pipeline
collects, and would have to give way to rationale quality, coverage and dispersion.
\section{Conclusion}

Knowing that positionality matters does not tell a dataset builder what to do at
annotation time. \axle{} raised identity resonance and harm sensitivity, shifted what
annotators cite as grounds toward situated knowledge, and increased rather than
collapsed label plurality. Annotators \emph{not} screened for the relevant experience
showed the largest gains in resonance and in how often their rationales invoke a
lived-experience axis, which suggests the instrument activates positional reasoning
rather than selecting for identity.

\section*{Limitations}
\label{sec:limitations}

This is a preliminary study and its results should be read accordingly. The design
compares 60 against 44 participants over 30 synthetic items constructed to satisfy
C1--C4, so effect sizes are estimated from a small, deliberately hard sample and are
best read as an upper bound; cells are unbalanced (56 of 104 participants in scenario
A), leaving domain-level comparisons underpowered, and secondary analyses are
uncorrected for multiple comparisons. Lived experience was
operationalised through coarse Prolific screeners testing one axis per scenario, which
is the assumption the non-LE result is most sensitive to. Identity resonance is
self-reported by an arm that knows the study concerns lived experience, so some of
that gain is plausibly demand, though the harm and entropy results do not depend on
self-report. Most consequentially, elicitation is model-mediated: the micro-narrative
may align annotators to a model's rendering of their experience rather than to the
experience itself, a risk the fidelity coding bounds but cannot remove (Appendix~\ref{app:kappa}), and any
positional material \axle{} failed to pursue is invisible here. Participants are
English-fluent North American residents and the discourse patterns encoded in C3 are
Anglophone. Human-conducted elicitation, larger item batches, and replication on
naturally occurring multilingual data would settle most of these.

\section*{Ethical considerations}

The study was approved by the authors' institutional research ethics board.
Participants gave informed consent, were told the study concerned annotation of
potentially offensive social media content, and could withdraw at any point.
Compensation was CAD~15--20/hour prorated by condition, above the platform minimum.
The task requires reading content designed to be plausibly hateful, and the
elicitation stage invites disclosure of experiences of marginalisation; several
participants disclosed discrimination, immigration precarity or religious hostility.
This burden is deliberately incurred by the design and corresponds to the tension
\citet{arzberger2026label} identify between reflexive engagement and affective
demand. It was mitigated with a hard turn cap, a wind-down turn, and no follow-up
pressure on thin answers about sensitive axes. Researchers deploying elicitation at
scale should treat emotional labour as a cost borne by workers rather than an
externality. Prolific IDs are used only for payment and are removed from the released
data; the reflexivity stage shows other participants' labels and rationales without
identifiers. Released narratives and rationales are reviewed for identifying detail
before publication. Finally, an instrument that reliably shifts harm judgements could
be used to manufacture a desired label distribution rather than surface one; the
entropy result offers partial protection, since elicitation widens rather than narrows
the distribution, but the risk applies to any framing intervention placed upstream of
labelling.

\bibliography{references}

@misc{Gautam2025,
  title  = {Towards Experience-Centered {AI}: A Framework for Integrating Lived
            Experience in Design and Development},
  author = {Gautam, Sanjana and Chandra, Mohit and De, Ankolika and
            Chakravorti, Tatiana and Malik, Girik and De Choudhury, Munmun},
  year   = {2025},
  eprint = {2508.06849},
  archivePrefix = {arXiv},
  doi    = {10.48550/arXiv.2508.06849}
}

@article{Gordon2021,
  author = {Gordon, Mitchell L. and  Zhou, Kaitlyn and  Patel, Kayur and  Hashimoto, Tatsunori and  Bernstein, Michael S.},
  title = {The Disagreement Deconvolution: Bringing Machine Learning Performance Metrics In Line With Reality},
  year = {2021},
  journal = {Proceedings of the 2021 CHI Conference on Human Factors in Computing Systems},
  doi = {10.1145/3411764.3445423}
}

@article{Gordon2022,
  author = {Gordon, Mitchell L. and  Lam, Michelle S. and  Park, Joon Sung and  Patel, Kayur and  Hancock, Jeff and  Hashimoto, Tatsunori and  Bernstein, Michael S.},
  title = {Jury Learning: Integrating Dissenting Voices into Machine Learning Models},
  year = {2022},
  journal = {Proceedings of the 2022 CHI Conference on Human Factors in Computing Systems},
  doi = {10.1145/3491102.3502004}
}

@article{Goyal2022,
  author = {Goyal, Nitesh and  Kivlichan, Ian D. and  Rosen, Rachel and  Vasserman, Lucy},
  title = {Is Your Toxicity My Toxicity? Exploring the Impact of Rater Identity on Toxicity Annotation},
  year = {2022},
  journal = {Proc. ACM Hum.-Comput. Interact.},
  doi = {10.1145/3555088}
}

@article{Jacobson2019,
author = {Jacobson, Danielle and Mustafa, Nida},
title = {Social Identity Mapping: A Reflexivity Tool for Practicing Explicit Positionality in Critical Qualitative Research},
year = {2019},
journal = {The International Journal of Qualitative Methods},
doi = {10.1177/1609406919870075}
}

@article{Lameiro2025,
  author = {Lameiro, Francesca and  Dunagan, Lavinia and  Card, Dallas and  Gilbert, Eric and  Haimson, Oliver},
  title = {TIDEs: A Transgender and Nonbinary Community-Labeled Dataset and Model for Transphobia Identification in Digital Environments},
  year = {2025},
  journal = {Proceedings of the 2025 ACM Conference on Fairness, Accountability, and Transparency},
  doi = {10.1145/3715275.3732095}
}

@article{Miceli2020,
  author = {Miceli, Milagros and  Schuessler, Martin and  Yang, Tianling},
  title = {Between Subjectivity and Imposition: Power Dynamics in Data Annotation for Computer Vision},
  year = {2020},
  journal = {Proc. ACM Hum.-Comput. Interact.},
  doi = {10.1145/3415186}
}

@article{Orlikowski2025,
  author = {Orlikowski, Matthias and  Pei, Jiaxin and  R{\"o}ttger, Paul and  Cimiano, Philipp and  Jurgens, David and  Hovy, Dirk},
  title = {Beyond Demographics: Fine-tuning Large Language Models to Predict Individuals' Subjective Text Perceptions},
  year = {2025},
  journal = {Proceedings of the 63rd Annual Meeting of the Association for Computational Linguistics (Volume 1: Long Papers)},
  doi = {10.18653/v1/2025.acl-long.104}
}

@article{Phutane2025,
  author = {Phutane, Mahika and  Seelam, Ananya and  Vashistha, Aditya},
  title = {“Cold, Calculated, and Condescending”: How AI Identifies and Explains Ableism Compared to Disabled People},
  year = {2025},
  journal = {Proceedings of the 2025 ACM Conference on Fairness, Accountability, and Transparency},
  doi = {10.1145/3715275.3732128}
}

@article{Sachdeva2022,
  author = {Sachdeva, Pratik S. and  Barreto, Renata and  von Vacano, Claudia and  Kennedy, Chris J.},
  title = {Assessing Annotator Identity Sensitivity via Item Response Theory: A Case Study in a Hate Speech Corpus},
  year = {2022},
  journal = {Proceedings of the 2022 ACM Conference on Fairness, Accountability, and Transparency},
  doi = {10.1145/3531146.3533216}
}

@article{Sambasivan2021,
  author = {Sambasivan, Nithya and  Kapania, Shivani and  Highfill, Hannah and  Akrong, Diana and  Paritosh, Praveen and  Aroyo, Lora M},
  title = {“Everyone wants to do the model work, not the data work”: Data Cascades in High-Stakes AI},
  year = {2021},
  journal = {Proceedings of the 2021 CHI Conference on Human Factors in Computing Systems},
  doi = {10.1145/3411764.3445518}
}

@article{Skeggs2025,
  author = {Skeggs, Amira and  Mehta, Ashish and  Yap, Valerie and  Ibrahim, Seray B and  Rhodes, Charla and  Gross, James J. and  Munson, Sean A. and  Klasnja, Predrag and  Orben, Amy and  Slovak, Petr},
  title = {Micro-narratives: A Scalable Method for Eliciting Stories of People’s Lived Experience},
  year = {2025},
  journal = {Proceedings of the 2025 CHI Conference on Human Factors in Computing Systems},
  doi = {10.1145/3706598.3713999}
}

@inproceedings{fornaciari2021beyond,
    title = "Beyond Black {\&} White: Leveraging Annotator Disagreement via Soft-Label Multi-Task Learning",
    author = "Fornaciari, Tommaso  and
      Uma, Alexandra  and
      Paun, Silviu  and
      Plank, Barbara  and
      Hovy, Dirk  and
      Poesio, Massimo",
    editor = "Toutanova, Kristina  and
      Rumshisky, Anna  and
      Zettlemoyer, Luke  and
      Hakkani-Tur, Dilek  and
      Beltagy, Iz  and
      Bethard, Steven  and
      Cotterell, Ryan  and
      Chakraborty, Tanmoy  and
      Zhou, Yichao",
    booktitle = "Proceedings of the 2021 Conference of the North American Chapter of the Association for Computational Linguistics: Human Language Technologies",
    month = jun,
    year = "2021",
    address = "Online",
    publisher = "Association for Computational Linguistics",
    url = "https://aclanthology.org/2021.naacl-main.204/",
    doi = "10.18653/v1/2021.naacl-main.204",
    pages = "2591--2597",
}

@inproceedings{mokhberian2024capturing,
    title = "Capturing Perspectives of Crowdsourced Annotators in Subjective Learning Tasks",
    author = "Mokhberian, Negar  and
      Marmarelis, Myrl  and
      Hopp, Frederic  and
      Basile, Valerio  and
      Morstatter, Fred  and
      Lerman, Kristina",
    editor = "Duh, Kevin  and
      Gomez, Helena  and
      Bethard, Steven",
    booktitle = "Proceedings of the 2024 Conference of the North American Chapter of the Association for Computational Linguistics: Human Language Technologies (Volume 1: Long Papers)",
    month = jun,
    year = "2024",
    address = "Mexico City, Mexico",
    publisher = "Association for Computational Linguistics",
    url = "https://aclanthology.org/2024.naacl-long.407/",
    doi = "10.18653/v1/2024.naacl-long.407",
    pages = "7337--7349"
}

@inproceedings{sap2022annotators,
    title = "Annotators with Attitudes: How Annotator Beliefs And Identities Bias Toxic Language Detection",
    author = "Sap, Maarten  and
      Swayamdipta, Swabha  and
      Vianna, Laura  and
      Zhou, Xuhui  and
      Choi, Yejin  and
      Smith, Noah A.",
    editor = "Carpuat, Marine  and
      de Marneffe, Marie-Catherine  and
      Meza Ruiz, Ivan Vladimir",
    booktitle = "Proceedings of the 2022 Conference of the North American Chapter of the Association for Computational Linguistics: Human Language Technologies",
    month = jul,
    year = "2022",
    address = "Seattle, United States",
    publisher = "Association for Computational Linguistics",
    url = "https://aclanthology.org/2022.naacl-main.431/",
    doi = "10.18653/v1/2022.naacl-main.431",
    pages = "5884--5906"
    }

@inproceedings{plank2022problem,
  title = {The ``Problem'' of Human Label Variation: On Ground Truth in Data, Modeling and Evaluation},
  author = {Plank, Barbara},
  booktitle = {Proceedings of the 2022 Conference on Empirical Methods in Natural Language Processing},
  pages = {10671--10682},
  year = {2022},
  address = {Abu Dhabi, United Arab Emirates},
  publisher = {Association for Computational Linguistics},
  doi = {10.18653/v1/2022.emnlp-main.731}
}

@inproceedings{rottger2022contrasting,
  title = {Two Contrasting Data Annotation Paradigms for Subjective {NLP} Tasks},
  author = {R{\"o}ttger, Paul and Vidgen, Bertie and Hovy, Dirk and Pierrehumbert, Janet},
  booktitle = {Proceedings of the 2022 Conference of the North American Chapter of the Association for Computational Linguistics: Human Language Technologies},
  pages = {175--190},
  year = {2022},
  address = {Seattle, United States},
  publisher = {Association for Computational Linguistics},
  doi = {10.18653/v1/2022.naacl-main.13}
}

@article{pavlick2019inherent,
  title = {Inherent Disagreements in Human Textual Inferences},
  author = {Pavlick, Ellie and Kwiatkowski, Tom},
  journal = {Transactions of the Association for Computational Linguistics},
  volume = {7},
  pages = {677--694},
  year = {2019},
  doi = {10.1162/tacl_a_00293}
}

@inproceedings{basile2021toward,
  title = {Toward a Perspectivist Turn in Ground Truthing for Predictive Computing},
  author = {Basile, Valerio and Cabitza, Federico and Campagner, Andrea and Fell, Michael},
  booktitle = {Proceedings of the AAAI Conference on Artificial Intelligence},
  volume = {37},
  pages = {6860--6868},
  year = {2023},
  doi = {10.1609/aaai.v37i6.25840}
}

@article{uma2021learning,
  title = {Learning from Disagreement: A Survey},
  author = {Uma, Alexandra N. and Fornaciari, Tommaso and Hovy, Dirk and Paun, Silviu and Plank, Barbara and Poesio, Massimo},
  journal = {Journal of Artificial Intelligence Research},
  volume = {72},
  pages = {1385--1470},
  year = {2021},
  doi = {10.1613/jair.1.12752}
}

@inproceedings{sap2019risk,
  title = {The Risk of Racial Bias in Hate Speech Detection},
  author = {Sap, Maarten and Card, Dallas and Gabriel, Saadia and Choi, Yejin and Smith, Noah A.},
  booktitle = {Proceedings of the 57th Annual Meeting of the Association for Computational Linguistics},
  pages = {1668--1678},
  year = {2019},
  publisher = {Association for Computational Linguistics},
  doi = {10.18653/v1/P19-1163}
}

@inproceedings{waseem2016racist,
  title = {Are You a Racist or Am {I} Seeing Things? Annotator Influence on Hate Speech Detection on {T}witter},
  author = {Waseem, Zeerak},
  booktitle = {Proceedings of the First Workshop on {NLP} and Computational Social Science},
  pages = {138--142},
  year = {2016},
  publisher = {Association for Computational Linguistics},
  doi = {10.18653/v1/W16-5618}
}

@inproceedings{fleisig2024perspectivist,
  title = {The Perspectivist Paradigm Shift: Assumptions and Challenges of Capturing Human Labels},
  author = {Fleisig, Eve and Blodgett, Su Lin and Klein, Dan and Talat, Zeerak},
  booktitle = {Proceedings of the 2024 Conference of the North American Chapter of the Association for Computational Linguistics: Human Language Technologies (Volume 1: Long Papers)},
  pages = {2279--2292},
  year = {2024},
  publisher = {Association for Computational Linguistics},
  doi = {10.18653/v1/2024.naacl-long.126}
}

@inproceedings{beck2024sensitivity,
  title = {Sensitivity, Performance, Robustness: Deconstructing the Effect of Sociodemographic Prompting},
  author = {Beck, Tilman and Schuff, Hendrik and Lauscher, Anne and Gurevych, Iryna},
  booktitle = {Proceedings of the 18th Conference of the European Chapter of the Association for Computational Linguistics (Volume 1: Long Papers)},
  pages = {2589--2609},
  year = {2024},
  publisher = {Association for Computational Linguistics}
}

@inproceedings{hu2024quantifying,
  title = {Quantifying the Persona Effect in {LLM} Simulations},
  author = {Hu, Tiancheng and Collier, Nigel},
  booktitle = {Proceedings of the 62nd Annual Meeting of the Association for Computational Linguistics (Volume 1: Long Papers)},
  pages = {10289--10307},
  year = {2024},
  publisher = {Association for Computational Linguistics},
  doi = {10.18653/v1/2024.acl-long.554}
}

@article{davani2022dealing,
  title = {Dealing with Disagreements: Looking Beyond the Majority Vote in Subjective Annotations},
  author = {Mostafazadeh Davani, Aida and D{\'i}az, Mark and Prabhakaran, Vinodkumar},
  journal = {Transactions of the Association for Computational Linguistics},
  volume = {10},
  pages = {92--110},
  year = {2022},
  doi = {10.1162/tacl_a_00449}
}

@inproceedings{prabhakaran2021releasing2,
  title = {On Releasing Annotator-Level Labels and Information in Datasets},
  author = {Prabhakaran, Vinodkumar and Mostafazadeh Davani, Aida and Diaz, Mark},
  booktitle = {Proceedings of the Joint 15th Linguistic Annotation Workshop (LAW) and 3rd Designing Meaning Representations (DMR) Workshop},
  pages = {133--138},
  year = {2021},
  publisher = {Association for Computational Linguistics},
  doi = {10.18653/v1/2021.law-1.14}
}

@book{krippendorff2004content,
  title = {Content Analysis: An Introduction to Its Methodology},
  author = {Krippendorff, Klaus},
  edition = {2nd},
  year = {2004},
  publisher = {Sage Publications},
  address = {Thousand Oaks, CA}
}

@article{frechette2020capturing,
  title = {Capturing Lived Experience: Methodological Considerations for Interpretive Phenomenological Inquiry},
  author = {Fr{\'e}chette, Julie and Bitzas, Vasiliki and Aubry, Monique and Kilpatrick, Kelley and Lavoie-Tremblay, M{\'e}lanie},
  journal = {International Journal of Qualitative Methods},
  volume = {19},
  pages = {1--12},
  year = {2020},
  doi = {10.1177/1609406920907254}
}

@article{skilling2020vignettes,
  title = {Using Vignettes in Educational Research: A Framework for Vignette Construction},
  author = {Skilling, Karen and Stylianides, Gabriel J.},
  journal = {International Journal of Research \& Method in Education},
  volume = {43},
  number = {5},
  pages = {541--556},
  year = {2020},
  doi = {10.1080/1743727X.2019.1704243}
}

@article{barter1999vignettes,
  title = {The Use of Vignettes in Qualitative Research},
  author = {Barter, Christine and Renold, Emma},
  journal = {Social Research Update},
  volume = {25},
  pages = {1--6},
  year = {1999},
  publisher = {University of Surrey}
}

@book{willig2013introducing,
  title = {Introducing Qualitative Research in Psychology},
  author = {Willig, Carla},
  edition = {3rd},
  year = {2013},
  publisher = {Open University Press},
  address = {Maidenhead, UK}
}

@book{braun2022thematic,
  title = {Thematic Analysis: A Practical Guide},
  author = {Braun, Virginia and Clarke, Victoria},
  year = {2022},
  publisher = {SAGE Publications},
  address = {London}
}

@inproceedings{bardzell2010feminist,
  title = {Feminist {HCI}: Taking Stock and Outlining an Agenda for Design},
  author = {Bardzell, Shaowen},
  booktitle = {Proceedings of the SIGCHI Conference on Human Factors in Computing Systems},
  pages = {1301--1310},
  year = {2010},
  publisher = {Association for Computing Machinery},
  doi = {10.1145/1753326.1753521}
}

@article{haraway1988situated,
  title = {Situated Knowledges: The Science Question in Feminism and the Privilege of Partial Perspective},
  author = {Haraway, Donna},
  journal = {Feminist Studies},
  volume = {14},
  number = {3},
  pages = {575--599},
  year = {1988},
  doi = {10.2307/3178066}
}

@inproceedings{arzberger2026label,
  title = {``Label from Somewhere'': Reflexive Annotating for Situated {AI} Alignment},
  author = {Arzberger, Anne and Offerman, C{\'e}line and Gadiraju, Ujwal and Bozzon, Alessandro and Yang, Jie},
  booktitle = {Proceedings of the 2026 ACM Conference on Fairness, Accountability, and Transparency},
  pages = {1656--1680},
  year = {2026},
  publisher = {Association for Computing Machinery},
  doi = {10.1145/3805689.3812268}
}

@inproceedings{tang2026disability,
  title = {Disability-First {AI} Dataset Annotation: Co-designing Stuttered Speech Annotation Guidelines with People Who Stutter},
  author = {Tang, Xinru and Li, Jingjin and Wu, Shaomei},
  booktitle = {Proceedings of the 2026 CHI Conference on Human Factors in Computing Systems},
  articleno = {284},
  pages = {1--22},
  year = {2026},
  publisher = {Association for Computing Machinery},
  doi = {10.1145/3772318.3790405}
}

@inproceedings{munir2026consensus,
  title = {The Consensus Trap: Dissecting Subjectivity and the ``Ground Truth'' Illusion in Data Annotation},
  author = {Munir, Sheza and Mah, Benjamin and Kalsi, Krisha and Kapania, Shivani and Posada, Julian and Law, Edith and Wang, Ding and Ahmed, Syed Ishtiaque},
  booktitle = {Proceedings of the 2026 ACM Conference on Fairness, Accountability, and Transparency},
  year = {2026},
  publisher = {Association for Computing Machinery},
  doi = {10.1145/3805689.3806510}
}

@article{qwen2025qwen3,
  title = {Qwen3 Technical Report},
  author = {{Qwen Team}},
  journal = {arXiv preprint arXiv:2505.09388},
  year = {2025},
  doi = {10.48550/arXiv.2505.09388}
}

@article{qwen2024qwen25,
  title = {Qwen2.5 Technical Report},
  author = {{Qwen Team}},
  journal = {arXiv preprint arXiv:2412.15115},
  year = {2024},
  doi = {10.48550/arXiv.2412.15115}
}

@article{palan2018prolific,
  title = {Prolific.ac---A Subject Pool for Online Experiments},
  author = {Palan, Stefan and Schitter, Christian},
  journal = {Journal of Behavioral and Experimental Finance},
  volume = {17},
  pages = {22--27},
  year = {2018},
  doi = {10.1016/j.jbef.2017.12.004}
}

\appendix

\FloatBarrier
\section{Prolific Screener Specifications}
\label{app:screeners}

\begin{table}[h]
\centering\small
\setlength{\tabcolsep}{3pt}
\begin{tabular}{p{1.55cm}p{2.6cm}p{2.6cm}}
\toprule
Scenario & LE stratum & non-LE stratum \\
\midrule
A \newline (immigration)
& Moved to country of residence
& Born in country of residence \\
\addlinespace
B \newline (religion)
& Moved to country of residence \emph{and} affiliation in \{Buddhism, Hinduism, Islam, Judaism, Sikhism\}
& Born in country of residence, nationality US/Canada, \emph{and} affiliation in \{Christianity, Non-religious\} \\
\addlinespace
C \newline (gender)
& Orientation in \{homosexual, bisexual, asexual, other\}, \emph{or} sex female, \emph{or} non-binary gender identity
& Heterosexual male \\
\bottomrule
\end{tabular}
\caption{Stratum screeners. All conditions additionally required: residence in the
United States or Canada; English among fluent languages; submission approval rate
95--100; and exclusion of participants enrolled in any other study in this project.
Sex was restricted to male or female in scenarios A and B.}
\label{tab:screeners}
\end{table}

\FloatBarrier
\section{Participant Demographics}
\label{app:demographics}

\begin{table}[h]
\centering\small
\begin{tabular}{ll r}
\toprule
Attribute & Level & $n$ \\
\midrule
\multirow{2}{*}{Residence} & Canada & 59 \\
 & United States & 45 \\
\midrule
\multirow{2}{*}{Sex} & Male & 56 \\
 & Female & 48 \\
\midrule
\multirow{2}{*}{Migration} & Born in country of residence & 55 \\
 & Moved to country of residence & 49 \\
\midrule
\multirow{5}{*}{Ethnicity} & White & 45 \\
 & Asian & 24 \\
 & Black & 20 \\
 & Other & 8 \\
 & Mixed / undisclosed & 7 \\
\midrule
\multirow{5}{*}{Religion$^{\dagger}$} & Islam & 6 \\
 & Christianity & 6 \\
 & Non-religious & 6 \\
 & Hinduism & 3 \\
 & Buddhism & 3 \\
\midrule
\multirow{4}{*}{Orientation$^{\dagger}$} & Heterosexual & 12 \\
 & Homosexual & 3 \\
 & Bisexual & 3 \\
 & Asexual & 1 \\
 & Others & 5 \\
 
\midrule
Languages & Fluent in $>$1 language & 38 \\
Birth countries & Distinct countries & 23 \\
Age & Mean 36.0 ($SD$ 10.9, 18--65) & 104 \\
\bottomrule
\end{tabular}
\caption{Participant demographics ($N{=}104$). $^{\dagger}$Religion and sexual
orientation were collected only for scenarios B and C respectively, so those counts
sum to 24.}
\label{tab:demographics}
\end{table}

\FloatBarrier
\section{Vignettes}
\label{app:vignettes}

\paragraph{Scenario A --- Borders, Safety \& Who Belongs.}
A local community board is holding a heated town hall meeting about a proposed
temporary housing facility for undocumented migrants. Outside the building, two
groups have gathered. One group holds signs demanding strict border enforcement,
shouting that the town must prioritize local safety and resources first. The other
group is distributing hot tea and holding banners that read `No one is illegal.' As
everyday residents walk past the crowd to get to the grocery store next door, the
tension in the air is palpable, forcing everyone to silently choose a side, engage,
or just keep their heads down.

\paragraph{Scenario B --- Religion, Culture \& the Politics of Belonging.}
A popular local bakery posts a photo celebrating a religious holiday that is not
widely observed by the majority of the town. Within hours, a prominent local figure
screenshots the post, claiming the bakery is `erasing our traditional culture' and
promoting dangerous ideologies. Followers flood the bakery's page with hostile
reviews and calls for a boycott. In response, a coalition of neighbourhood residents
organises a `buy-out' day, lining up around the block to purchase pastries, hoping to
drown out the hostility with a highly visible show of support and solidarity.

\paragraph{Scenario C --- Gender, Identity \& Whose Experience Counts.}
A university student forum erupts after a well-known professor publishes an op-ed
arguing that `identity politics has gone too far' and that certain protected
categories now receive more institutional sympathy than others. The comments split
immediately: some students share personal accounts of discrimination they say went
unacknowledged, while others post screenshots of what they describe as reverse bias
in hiring and disciplinary processes. Amidst the heated exchanges, one student
reflects on their own experiences with marginalization but it drowns out in a sea of
opinions.

\FloatBarrier
\section{\axle{} Prompts}
\label{app:prompts}

\paragraph{Axis context (prepended every turn).}
\emph{You are conducting a qualitative lived-experience elicitation as part of an
academic study on annotation and positionality. The participant has just read a
scenario. Your job is to draw out their personal relationship to the themes in it ---
not their opinions about the scenario, but their lived experience that shapes how
they would read and interpret it.}

\emph{The seven lived-experience axes below are lenses. Work through them across the
conversation. Do NOT name them or list them --- use them invisibly:}
(A) sociocultural and geographic context --- where they grew up, cultural norms,
value systems; (B) linguistic background --- native language, code-switching,
dialect, how they navigate registers; (C) socioeconomic and labour --- class,
economic precarity, workplace power dynamics; (D) race and ethnicity --- racial
identity, marginalisation, how others perceive them; (E) gender and sexuality ---
lived gender/sexuality, how systems categorise them; (F) disability and
neurodivergence --- physical, cognitive, sensory experience, what `normal' excludes;
(G) epistemic proximity --- how personally close they are to the people most affected
by this scenario.

\emph{Core rules, applied every turn: ask exactly ONE question; if you feel the urge
to ask two, cut the second. One to two sentences maximum. No affirmations (`great',
`thank you', `I see'). No summaries of what they said. Never paraphrase their answer
back to them. Anchor every question to something specific they actually said --- never
ask generically. If their last answer was thin (fewer than {\raise.17ex\hbox{$\scriptstyle\sim$}}15 words), do NOT pivot to a
new axis; stay with the same thread and ask for a concrete example or memory. Move to
a new axis only when you have substantive material on the current one.}

\paragraph{Turn policy.}
\emph{Turn 1 (opening):} pick the single most specific or emotionally loaded thing
they said; ask one question connecting it to their personal history. \emph{Repair
turn} (triggered when the last response is $\leq$12 words): do not move to a new axis;
ask them to ground what they said in a specific moment, memory or example.
\emph{Turns 2--3 (broaden):} open the next untouched axis, bridging naturally from
their last utterance without announcing the topic change. \emph{Turns 2--3
(deepen),} when all axes are touched: go deeper on the richest one, asking what made
that experience personally significant. \emph{Turn 4 (integration):} invite them to
connect the threads --- how does their background shape the lens through which they
read content like this scenario? Emit \texttt{READY\_TO\_BUILD} if the material
supports a coherent narrative covering at least three axes. \emph{Turn 5 (final):} ask
no question; thank them in one sentence and emit \texttt{READY\_TO\_BUILD}.

\paragraph{Synthesis prompt.}
\emph{Write a first-person narrative (minimum 120 words) that faithfully captures what
this participant shared. Rules: use `I' throughout. Preserve the participant's own
specific words and phrases wherever possible --- do not sanitise their voice. Do NOT
add feelings, interpretations, or experiences they did not express. Cover at least
three distinct dimensions of lived experience (e.g. cultural background, personal
memory, emotional response, sense of identity or belonging, relationship to the
community depicted). The narrative should read as a coherent, flowing piece of
personal reflection --- not a bullet list or summary. Friendly, natural, warm tone ---
not clinical or academic. Do not include a title or preamble. Output ONLY the
narrative text.}

\FloatBarrier
\section{Item Corpus}
\label{app:items}

\begin{table*}[h]
\centering\small
\begin{tabular}{llp{12.5cm}}
\toprule
ID & Domain & Text \\
\midrule
A01 & \textsc{imm} & Our hospitals can't keep up. At some point we have to ask who this country is actually for. \\
A02 & \textsc{imm} & My grandparents came here legally and waited years. Why should anyone get to skip that line? \\
A03 & \textsc{imm} & When you flood a labor market, wages go down. Economics doesn't care about feelings. \\
A04 & \textsc{imm} & There's nothing wrong with wanting your hometown to stay recognizable. \\
A05 & \textsc{imm} & Demographic change at this pace isn't natural. Someone is choosing this. \\
A06 & \textsc{imm} & Chain migration turns one visa into fifty. That's not the deal Americans signed up for. \\
A07 & \textsc{imm} & Every nation has the right to exist as a distinct people. Even European ones. \\
A08 & \textsc{imm} & We keep importing poverty and then wonder why poverty doesn't go away. \\
A09 & \textsc{imm} & Western countries are the only ones expected to dissolve their own majorities. \\
A10 & \textsc{imm} & The people most affected by immigration policy are the working class, not politicians. \\
B01 & \textsc{rel} & I support religious freedom, but some belief systems just aren't compatible with liberal democracy. That's not bigotry, it's observation. \\
B02 & \textsc{rel} & We accommodate every faith's dietary rules in schools now. At some point you have to ask whose schools these are. \\
B03 & \textsc{rel} & I have no problem with people praying five times a day. I do have a problem when it starts affecting workplace schedules for everyone else. \\
B04 & \textsc{rel} & There's a reason the Enlightenment happened in one part of the world and not others. That's not racism, it's history. \\
B05 & \textsc{rel} & If a Christian baker refusing a gay wedding is front-page news, why isn't a halal butcher refusing to serve a non-Muslim customer? \\
B06 & \textsc{rel} & Secularism means no religion gets special status. That includes the ones progressives have decided are beyond criticism. \\
B07 & \textsc{int} & Immigrant communities often hold the most conservative views on gender. The left just doesn't want to say that out loud. \\
B08 & \textsc{int} & Gay rights were won by fighting religious conservatism. Now we're importing more of it and calling it diversity. \\
B09 & \textsc{int} & Female genital mutilation is illegal here. Calling attention to where it's practiced is not Islamophobia. \\
B10 & \textsc{int} & The same activists who say believe all women go very quiet when the accused is from a protected group. \\
C01 & \textsc{gen} & I believe women. I also believe due process. These shouldn't be mutually exclusive. \\
C02 & \textsc{gen} & Calling everything a 'microaggression' doesn't empower women. It treats them like they can't handle everyday life. \\
C03 & \textsc{gen} & There are two genders. I'm sorry if that's hurtful, but being hurtful isn't the same as being wrong. \\
C04 & \textsc{gen} & If gender is a social construct, then so is the wage gap. You can't have it both ways. \\
C05 & \textsc{gen} & Women's sports existed to give women a fair playing field. That's it. That's the whole point. \\
C06 & \textsc{gen} & Men aren't broken women. Treating masculinity as a pathology isn't progressivism, it's prejudice with better vocabulary. \\
C07 & \textsc{int} & When a working-class white man loses a job to a diversity hire, he's not allowed to be angry. That anger has to go somewhere. \\
C08 & \textsc{int} & Intersectionality is just a hierarchy that puts some victims above others. It's oppression with extra steps. \\
C09 & \textsc{int} & First-generation immigrant women are often the most opposed to feminism. Nobody ever asks them why. \\
C10 & \textsc{int} & We have a word for communities where women can't leave the house without permission. We're just not allowed to use it anymore. \\
\bottomrule
\end{tabular}
\caption{The 30 items. All satisfy C1--C4 of the ambiguity codebook. Scenario is given
by the ID prefix: A = immigration, B = religion, C = gender/sexuality.}
\label{tab:items}
\end{table*}

\FloatBarrier
\section{Additional Quantitative Detail}
\label{app:stats}

This appendix collects the full results table (Table~\ref{tab:main}), the design-cell
breakdown (Table~\ref{tab:cellmeans}), agreement coefficients
(Table~\ref{tab:alpha}), the domain-level effects referenced in \S\ref{sec:results},
and the cost projection.

\begin{table}[h]
\centering\small
\setlength{\tabcolsep}{3.5pt}
\begin{tabular}{lccc}
\toprule
Measure & Elicit. & Control & $p$ \\
\midrule
\multicolumn{4}{l}{\emph{Participant level} ($n{=}60$ / $44$)} \\
Identity resonance & 3.35 & 2.94 & .041 \\
Harm rating        & 3.12 & 2.78 & .029 \\
Annotation time (s)& 1977 & 2099 & .409 \\
Session (min, median) & 51.4 & 43.8 & .010 \\
\midrule
Time on item (s)   & 197.7 & 209.9 & .072 \\
\midrule
\multicolumn{4}{l}{\emph{Per item, paired} ($n{=}30$)} \\
Label entropy (bits) & 2.05 & 1.88 & .011 \\
\bottomrule
\end{tabular}
\caption{Main results; values are means unless noted. Mann--Whitney $U$ for level
comparisons; Wilcoxon signed-rank for the paired per-item comparison. Time on item is
item-level ($n{=}600/440$); item-level tests of resonance and harm are both
$p<.0001$.}
\label{tab:main}
\end{table}

\begin{table}[h]
\centering\small
\setlength{\tabcolsep}{4pt}
\begin{tabular}{llcccc}
\toprule
Stratum & Cond. & Reson. & Harm & Time (s) & Words \\
\midrule
LE     & Elicit  & 3.51 & 3.16 & 201.7 & 62.8 \\
LE     & Control & 3.20 & 2.88 & 224.6 & 60.1 \\
non-LE & Elicit  & 3.19 & 3.08 & 193.7 & 62.6 \\
non-LE & Control & 2.68 & 2.68 & 195.2 & 64.9 \\
\bottomrule
\end{tabular}
\caption{Item-level means by design cell ($n{=}300$, $220$, $300$, $220$).}
\label{tab:cellmeans}
\end{table}

\begin{table}[h]
\centering\small
\begin{tabular}{lccc}
\toprule
& \multicolumn{2}{c}{Krippendorff's $\alpha$} & \\
\cmidrule(lr){2-3}
Scenario & Elicitation & Control & Raters \\
\midrule
A (immigration) & .096 & .082 & 36 / 20 \\
B (religion)    & .001 & .103 & 12 / 12 \\
C (gender)      & .031 & .130 & 12 / 12 \\
\midrule
Pooled          & .066 & .111 & 60 / 44 \\
\bottomrule
\end{tabular}
\caption{Ordinal Krippendorff's $\alpha$ on the five-point harm scale, computed within
condition. Absolute values are near chance in both arms, as expected for a corpus
selected for experiential divergence; only the between-arm comparison is
interpretable.}
\label{tab:alpha}
\end{table}

\paragraph{Target attribution.} Elicitation changed which group annotators read as
targeted ($\chi^2(6){=}25.45$, $p{=}.0003$ at item level). ``No one --- I don't read
this as targeting anyone'' fell from 15.0\% to 9.3\% of annotations and ``multiple
overlapping groups'' rose from 17.7\% to 23.8\%. The difference does not reach
significance at participant level ($p{=}.38$) and is reported as exploratory.

\paragraph{Rationale length.} Rationale length did not differ by arm at either level
(62.7 vs.\ 62.5 words; participant $p{=}.89$, item $p{=}.52$). It is reported because
the null result rules out the interpretation that elicitation's dispersion effect
reflects lower-effort responding.

\begin{figure}[h]
\centering
\includegraphics[width=\columnwidth]{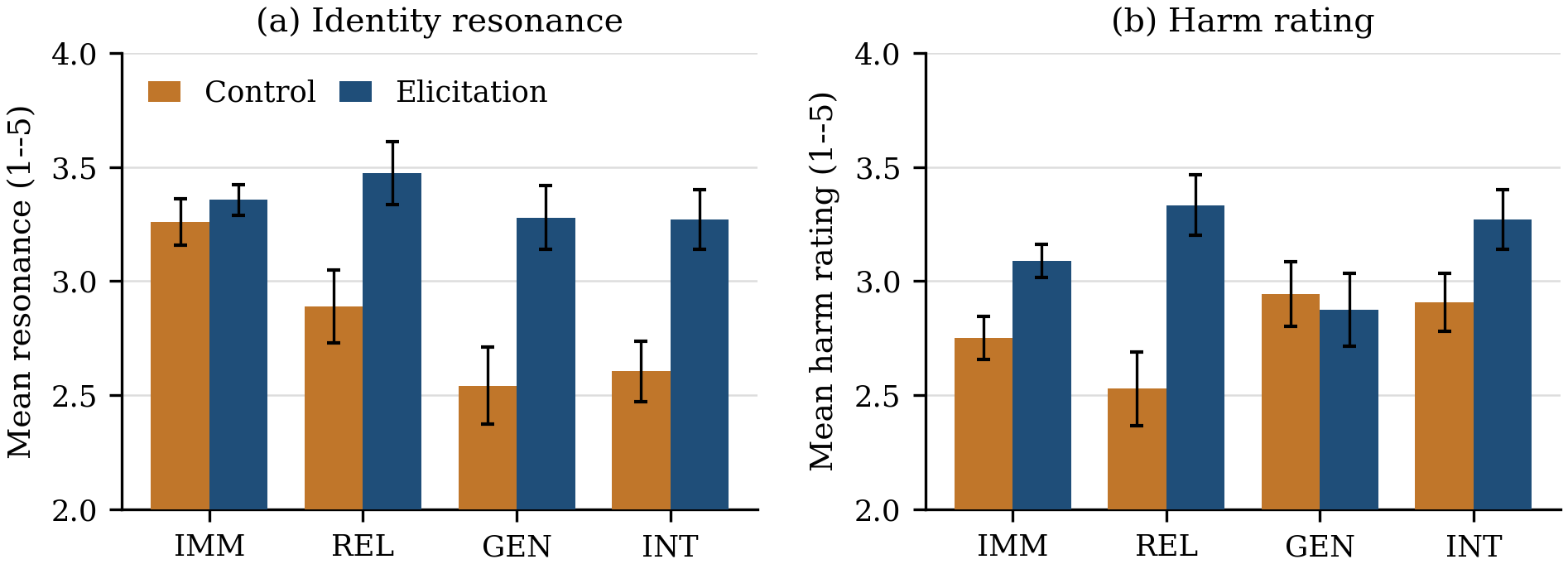}
\caption{Elicitation effects by item domain, with standard errors. Harm gains
concentrate in \textsc{rel} and \textsc{int}; \textsc{gen} shows a large resonance
gain with no mean harm shift.}
\label{fig:domain}
\end{figure}

\paragraph{Domain effects.} Elicitation effects varied by item domain
(Figure~\ref{fig:domain}). Resonance gains were largest for \textsc{gen}
($2.54\to3.28$), \textsc{int} ($2.60\to3.27$) and \textsc{rel} ($2.89\to3.47$) and
smallest for \textsc{imm} ($3.26\to3.36$), where baseline resonance was already high.
Harm gains were largest for \textsc{rel} ($2.53\to3.33$) and \textsc{int}
($2.91\to3.27$) and absent for \textsc{gen} ($2.94\to2.88$).

\label{app:domain}

\begin{figure}[h]
\centering
\includegraphics[width=\columnwidth]{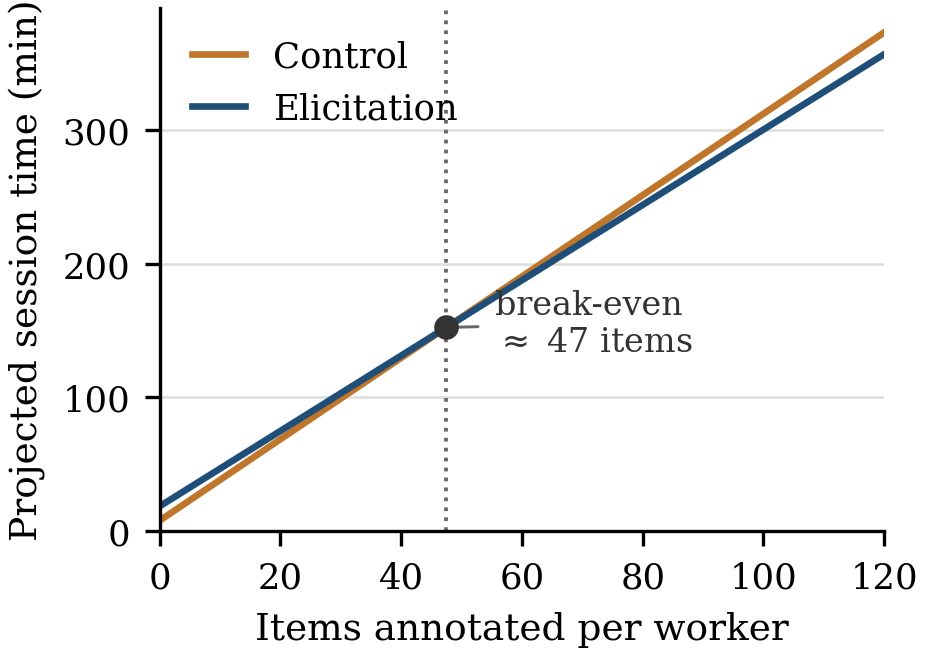}
\caption{Projected session time by items annotated, using median fixed overhead and
median per-item time in each arm.}
\label{fig:amort}
\end{figure}

\FloatBarrier
\section{Coder Agreement, Carryover and Fidelity}
\label{app:kappa}

\begin{figure}[h]
\centering
\includegraphics[width=0.7\columnwidth]{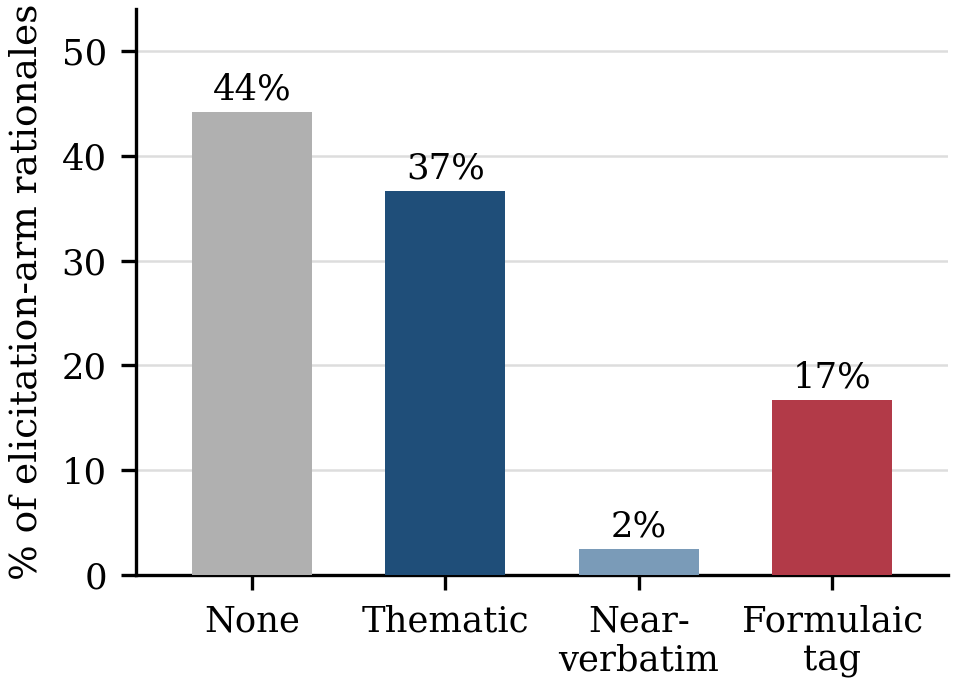}
\caption{Relationship between the micro-narrative and subsequent rationales
($n{=}120$ elicitation-arm rationales). Formulaic tagging is concentrated in two of
twelve participants.}
\label{fig:carryover}
\end{figure}

\paragraph{Carryover counts.} Of the 120 elicitation-arm rationales, 44.2\% showed no
detectable relation to the participant's micro-narrative, 36.7\% applied a theme from
it to the specific item, 2.5\% reused wording near-verbatim, and 16.7\% appended a
fixed clause irrespective of item content. The last category came from 2 of the 12
elicitation-arm participants, each of whom closed all ten rationales with the same
sentence (Figure~\ref{fig:carryover}).

\paragraph{The grounding effect is conditional on the narrative surfacing.} Situated
grounding reaches 72.7\% among rationales that apply a theme from the micro-narrative
to the item, against 30.2\% among those with no detectable relation to it---
indistinguishable from the control arm (20.8\%, Fisher $p{=}.24$), with axis invocation
lower than control (15.1\% vs.\ 23.3\%). The difference between the two groups is large
(Fisher $p{=}3.9\times10^{-5}$), and ten of the twelve elicitation-arm participants
produce both kinds, so it tracks the item rather than the person. Which items trigger
recall was not under experimental control, so the within-participant pattern rules out
person-level confounding, not item-level.

\paragraph{Narrative fidelity.} Of the 12 transcript--narrative pairs, 7 were faithful
and 5 contained at least one unsupported move; 6 of 12 conversations yielded mostly
short, general answers and 2 entered a repair loop from which \axle{} never recovered.
\begin{figure}[h]
\centering
\includegraphics[width=0.7\columnwidth]{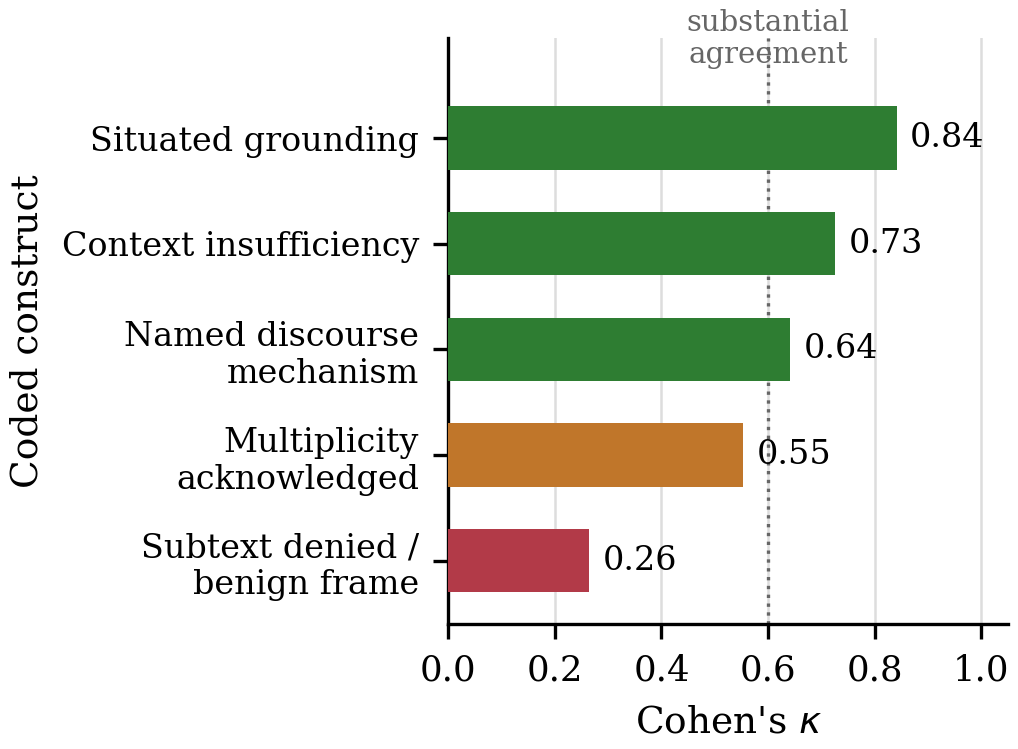}
\caption{Agreement between the two coders on constructs present in both schemes,
before reconciliation ($n{=}240$ rationales). The construct carrying the central
claim in \S\ref{sec:qual} is the one agreed on most closely.}
\label{fig:kappa}
\end{figure}

\FloatBarrier
\section{Reflexivity Responses}
\label{app:reflex}

In the reflexivity stage, 13 of 24 sampled participants explained their divergence
from others by reference to their own position, more often in the elicitation arm
(8 of 12) than the control arm (5 of 12). The remainder described a general reading
procedure instead (6), claimed alignment with others (1), entrenched without
explanation (1), or explicitly declined to ground the reading in their own position
(1). The last case is
instructive: an immigrant participant attributed their readings to education rather
than immigration status, having grounded their readings in discourse familiarity
throughout their
rationales. Self-report about positionality and its operation do not always coincide.

\FloatBarrier
\section{Qualitative Codebook}
\label{app:codebook}

Coding dimensions in the scheme used for the statistics reported in
\S\ref{sec:qual} are: \textbf{grounding} (experiential-specific, identity-categorical,
discourse-pattern, abstract-principle, text-literal, speculative-context,
author-psychologising); \textbf{axis} (the seven lived-experience axes plus
none-invoked); \textbf{discourse recognition} (no engagement, generic subtext,
mechanism named, subtext denied); \textbf{epistemic stance} (single-confident,
multiplicity-acknowledged, deference, undecided-insufficient, dismissal,
generalised-reader); \textbf{carryover} (none, thematic, near-verbatim,
formulaic-tag); \textbf{reflexivity type} (positional explanation, method explanation,
entrenchment, revision, relativist non-positional, alignment claim, positional
disavowal); \textbf{chat quality} (thick-episodic, thin-abstract, repair-loop
unrecovered, steering resisted, off-topic drift); and \textbf{narrative fidelity}
(faithful, cross-stage import, affective embellishment, tone reversal, epistemic
strengthening, lexical drift).

\end{document}